\documentclass{article}

\usepackage{spconf,amsmath,epsfig, cite}
\usepackage{graphicx}
\usepackage{amsmath}
\usepackage{amssymb}
\usepackage{subfig}
\usepackage{multirow}
\usepackage{dsfont}
\usepackage{colortbl}
\usepackage[table]{xcolor}
\usepackage{tabu}

\usepackage{etoolbox}
\patchcmd{\thebibliography}
  {\settowidth}
  {\setlength{\itemsep}{0pt plus 0.1pt}\settowidth}
  {}{}
\apptocmd{\thebibliography}
  {\small}
  {}{}

\newcommand{\etal}{\textit{et al}.}
\definecolor{TableRed}{HTML}{800000}
\newcommand{\TextRed}[1]{\textcolor{TableRed}{#1}}
\title{Angular Luminance for Material Segmentation}
\name{Jia Xue, Matthew Purri, Kristin Dana}
\address{Department of Electrical and Computer Engineering, Rutgers University, Piscataway, NJ 08854}
\begin{document}
%
\maketitle
\begin{abstract}
Moving cameras provide multiple intensity measurements per pixel, yet often semantic segmentation, material recognition, and object recognition do not utilize this information. With basic alignment over several frames of a moving camera sequence, a distribution of intensities over multiple angles is obtained. It is well known from prior work that luminance histograms  and the statistics of natural images provide a strong material recognition cue. We utilize per-pixel {\it angular luminance distributions} as a key feature in discriminating the material of the surface. The angle-space sampling in a multiview satellite image sequence is an unstructured sampling of the underlying reflectance function of the material. For real-world materials there is significant intra-class variation  that can be managed by building a angular luminance  network (AngLNet). This network combines angular reflectance cues from multiple images with  spatial cues  as input to fully convolutional networks for material segmentation.
We demonstrate the increased performance  of AngLNet over prior state-of-the-art in material segmentation from  
satellite imagery.
\end{abstract}
\vspace{-0.15in}
\section{Introduction}
\vspace{-0.08in}
Material recognition and segmentation is an important area of research in computer vision for providing more complete scene understanding. 
Material segmentation assigns a material id  to every pixel in an image based on texture and reflectance cues.  Semantic segmentation  assigns an object class to every image pixel based on shape, color and position cues as  well as global context information.   To underscore the difference between material and object-based semantic segmentation, consider that the same semantic object can be made of  different materials. For example, a building rooftop may be made of wood,  metal, concrete, polymer, or asphalt. 

Materials can often be distinguished using the bidirectional reflectance distribution function (BRDF) as done in 
traditional methods that use a full BRDF  \cite{dana1999reflectance, weyrich2006analysis} 
or  dense partial samplings of the BRDF  \cite{zhang2015reflectance, liu2014discriminative}
to characterize a material.
Capturing a full BRDF is rarely practical in mobile applications. However, multiview observations of a scene, even with irregular angular sampling,    provides an opportunity to use reflectance for recognition, since aligned imagery  gives a vector of reflectance samples per pixel.

Early studies in material recognition from reflectance characteristics largely concentrated on per-image recognition, which predicts one material class for the entire image or region. But pixel-wise prediction is required for segmentation, and characterizing material appearance with a BRDF for each pixel is an insurmountable task. Also, in many cases, we lack sufficient training data to formulate a probability distribution over the entire space of realizable BRDFs that fully capture the intraclass variation. 

In real-world image sequences the view/illumination angles are not fixed. The observation angles and observing sequences are different for different samples. The approach by Dror \etal \cite{dror2001estimating} asserts that given a finite but arbitrary set of candidate reflectance functions, we can identify which one most closely represents an observed surface by a low-cost intensity distribution. The method of luminance histograms used this idea and successfully characterized appearance by natural image statistics \cite{motoyoshi2007image, sharan2008image, zhu2016generative}.

Inspired by this,
we make use of a per-pixel {\it angular luminance histogram} representing the distribution of intensities observed per-pixel from all viewing angles in the multiview sequence. 
Integrated in a meaningful way within a large deep network, the angular histogram provides a cue that consistently improves performance for material-based segmentation.  Moreover, in applications such as satellite imaging, where multiview images are collected,  the angular histogram is readily available with little additional cost.

\vspace{-0.15in}
\section{Angular Histogram}
\vspace{-0.08in}

\begin{figure*}[t]
\centering
\includegraphics[width=15cm, height=10cm, keepaspectratio]{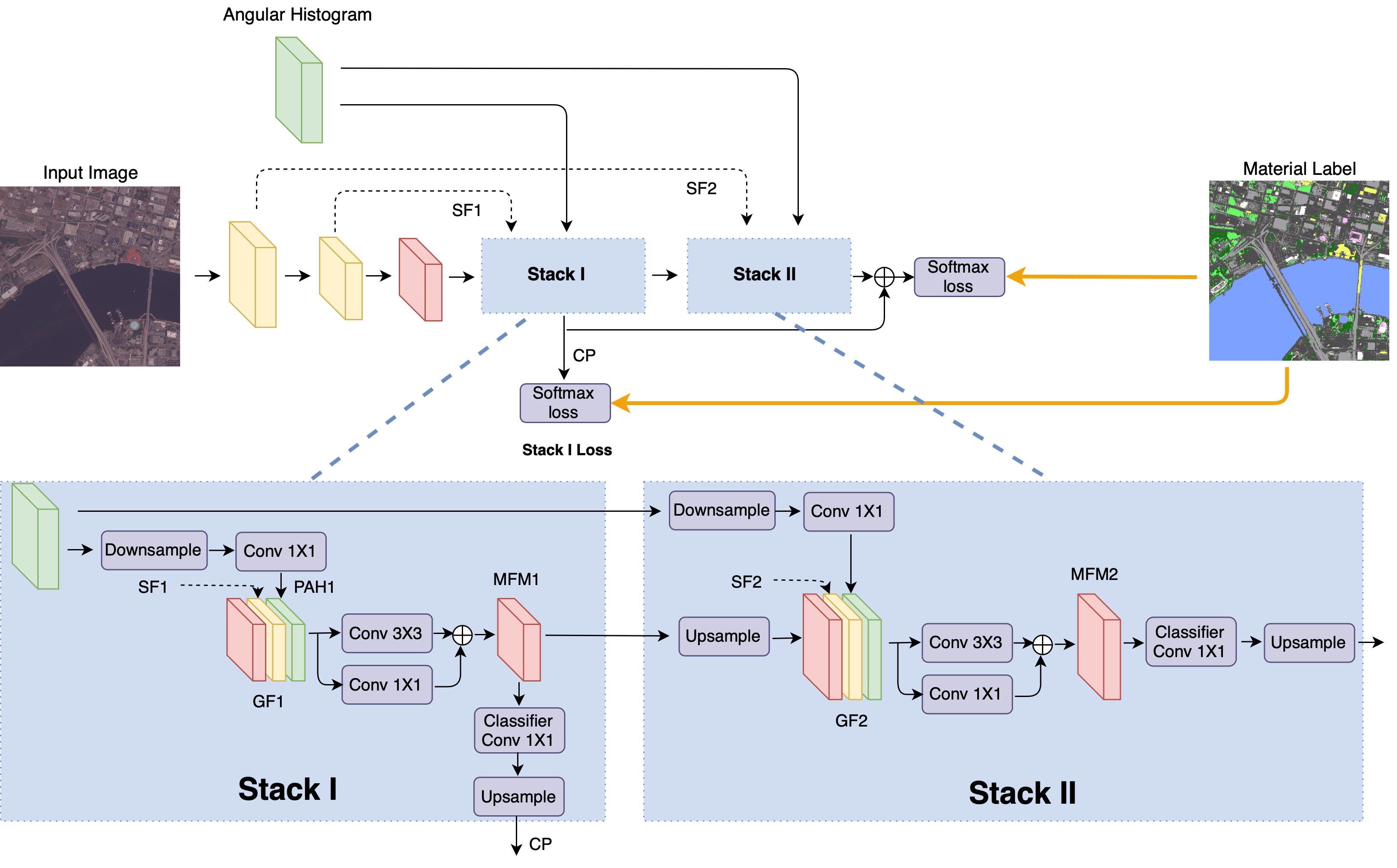}
\caption{Overall architecture of the proposed AngLNet. AngLNet makes prediction with a coarse-to-fine stacking network, where stack I learns the coarse prediction, and stack II learns the residual for refinement. (Notation: $\bigoplus$ means element-wise addition, CP means coarse prediction, GF means group features, SF1 and SF2 are the shallow layer features, MFM1 and MFM2 are the Material Feature Maps.)}
\label{fig:algorithm}
\vspace{-0.15in}
\end{figure*}

The bidirectional reflectance distribution function (BRDF) specifies how much of the light incident is reflected by an opaque surface patch in each possible view direction. BRDF is a function of four continuous angular variables. Let $l$ be the light direction, $v$ represents the view direction, the BRDF function is expressed as ${f}({\theta}_{i},{\varphi}_{i};{\theta}_{r},{\varphi}_{r})$, where ${\theta}_{i}$, ${\theta}_{r}$ and ${\varphi}_{i}$, ${\varphi}_{r}$ are the polar and azimuth angles of $I$ and $v$ respectively. If $I({\theta}_{i},{\varphi}_{i})$ represents the illumination incident, the total reflected radiance $I({\theta}_{r},{\varphi}_{r})$ of the surface patch to the view direction is 
\begin{equation}
      \int_{{\varphi}_{i}=0}^{2\pi}  \int_{{\theta}_{i}=0}^{\pi/2} f({\theta}_{i},{\varphi}_{i};{\theta}_{r},{\varphi}_{r})I({\theta}_{i},{\varphi}_{i})\cos{{\theta}_{i}}\sin{{\theta}_{i}} \,d{{\theta}_{i}}d{{\varphi}_{i}}.
\end{equation}
Replacing ${\theta}_{i}$, ${\varphi}_{i}$ and ${\theta}_{r}$, ${\varphi}_{r}$ with $l$ and $v$, the BRDF function becomes ${f}_{r}(l,v)$, and the relationship between the reflected radiance ${I}_{o}(v)$ and the illumination incident ${I}_{i}(l)$ is
\begin{equation}
    {I}_{o}(v) = {I}_{i}(l){f}_{r}(l,v)
\end{equation}
For different material classes $c=1,2 \dots k$, the BRDF function is described as ${f}_{cr}(l,v)$. With the same input light ${I}_{i}(l)$, the observed output light is  determined by the material BRDF function ${f}_{cr}(l,v)$. The BRDF is a unique identifier of the materials. In situations where the lighting condition is unknown and distributed similarly across the scene, the reflective distribution of each material is altered but the relative distribution between materials remains the same. Although we lack enough information to formulate the BRDF functions, we can identify which one most closely represents an observed surface with a finite set of candidate reflectance functions. In our dataset, we have 10 material classes, which corresponds to 10 reflectance functions.

Histogram distribution for material BRDF representation has been explored in many previous works \cite{dror2001estimating}.
But many focus on material recognition or 3D rendering, where the single image only contains single material. For material segmentation, the single image contains multiple different materials, so a per-pixel material id is required. For efficient computation, we build the angular histogram per-superpixel instead of per-pixel.
In our experiments, we employ SLIC \cite{achanta2012slic} as the superpixel algorithm and set the number of superpixels be 2000 for each image. 
For each superpixel segmented material patch, we compute the histogram distribution from a stack of multiview material patches, which are at the same spatial location of the multiview images. Observing that most of the pixel intensities are concentrated on a relatively small  region, for the setting of histogram bins, we use a multi-scale bin setting. We set a relatively large bin for the whole range of pixel values, and set another relatively small bin for the narrow concentrated pixel value range. Finally, we concatenate the multi-scale histogram intensities as our angular histogram. 

\begin{table*}[h]
\centering
\begin{tabu}{cccc|c|c|c|c}
\tabucline[1pt]{-}
\multirow{2}{*}{Method} &\multirow{2}{*}{BaseNet} & \multirow{2}{*}{\begin{tabular}[c]{@{}c@{}}Angular \\ Histogram\end{tabular}} & \multirow{2}{*}{\begin{tabular}[c]{@{}c@{}}Stacking\\ Network\end{tabular}} & \multicolumn{2}{c}{Jacksonville} & \multicolumn{2}{c}{Omaha} \\
 &  & & & pixACC & mIoU & pixACC & mIoU \\
 \hline
 \hline
Baseline & ResNet18 &  &  & 72.6 & 26.6 & 66.9 & 30.8 \\
AngLNet & ResNet18 & \checkmark & & 73.9 & 28.5 & 68.3 & 31.7\\
AngLNet & ResNet18 & \checkmark & \checkmark & 74.7 & 28.9 & 68.9 & 31.9  \\
AngLNet & ResNet50 & \checkmark & \checkmark & 75.2 & 29.9 & 68.5 & 30.6  \\
\tabucline[1pt]{-}
\end{tabu}
\caption{Results comparing performance with different components of AngLNet for material segmentation. AngLNet + Angular histogram means concatenating angular luminance histogram with features from dilated FCN baseline for segmentation, i.e.\ without the Stack II process in Figure~\ref{fig:algorithm}. Stacking Network means prediction from both Stack I and Stack II, the final prediction is the addition of coarse prediction from Stack I with refinement prediction from Stack II.}
\label{tab:ablation}
\vspace{-0.1in}
\end{table*}

\vspace{-0.15in}
\section{Angular Luminance Network}
\vspace{-0.08in}
We develop a new architecture, the Angular Luminance Network (AngLNet),
incorporating the angular luminance histogram and a novel configuration of network elements that encompasses recent developments in deep learning. 
The  angular histograms provide a $ h \times w \times b$ feature (where $b$ is the bin size) that is integrated with  pre-trained CNNs in a coarse-to-fine manner 
as illustrated in Figure~\ref{fig:algorithm}. Following prior work \cite{zhang2018context},  we use a pre-trained ResNet \cite{he2016deep} model with dilated network strategy 
to extract feature maps. The angular histograms are downsampled to the same height/width as the extracted feature maps of the ResNet architecture through a sequence of $1\times1$ convolutional and non-linear mapping layers. 
As shown in \cite{zeiler2014visualizing}, neurons from shallow layers become stimulated by information resembling texture. We use skip connections to propagate low level features for material segmentation. Features from shallow layers (labelled SF1 in Figure~\ref{fig:algorithm}), features from last residual block, and projected angular histograms (PAH1) are concatenated as group features (GF). The group features go through a newly designed residual block and generate the  Material Feature Map  (MFM1). Generating this MFM1 is the process labelled Stack I in Figure~\ref{fig:algorithm}. The MFM1 is fed in two directions: in one direction the MFM1 is upsampled by 2 and fed into the Stack II process where it is combined with the projected angular histogram (PAH2) and the shallow layer feature maps (SF2). 
The other direction generates a coarse prediction (CP). Based on the observation in  \cite{chen2017rethinking}, that upsampling the prediction for loss computation gives better performance, we keep the ground truth at the original resolution and upsample the prediction to the same size of ground truth for the coarse prediction loss computation. To optimize the learning process and fully utilize features learned from Stack I stage (MFM1), we use a coarse-to-fine residual learning that 
predicts the fine segmentation  by the addition of the CP and prediction from Stack II.
In this way, Stack II prediction is only responsible for residual refinement learning and the overall network is learned in a coarse-to-fine manner. 
In network training, the computed loss is a combination of the fine prediction loss and the coarse prediction loss multiplied by a loss weight.
In network testing, the accuracy and mIoU are calculated based on the fine  prediction only.

Stacking networks has been applied in many different areas \cite{ilg2017,zhang2017stackgan}.
The advantage of stacking networks is to mitigate the single network burden for complex tasks. The first network in the stack can generate a coarse prediction, and the second network is responsible for refinement. We use a two stages coarse-to-fine network for material segmentation, as shown in Figure~\ref{fig:algorithm}, we combine the angular histogram with features extracted from pre-trained CNN in two network stages. Prediction in Stack I comes from small size height/width  feature maps; it is responsible for coarse prediction. Prediction in Stack II comes from upsampled feature maps, and it is responsible for refinement. To optimize the learning process and fully leverage the residual learning strategy, the final prediction is the addition of prediction from Stack I and prediction from Stack II. Our comparison results in Section~\ref{sec:experiments} indicate the importance of coarse-to-fine residual learning.

\vspace{-0.25in}
\section{Experiments}
\label{sec:experiments}
\vspace{-0.08in}

\begin{figure*}[t]
\centering
\subfloat
{
\includegraphics[width=.13\linewidth]{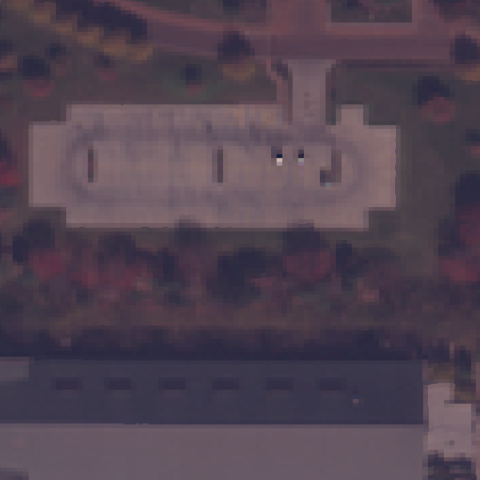}
}
\subfloat
{
\includegraphics[width=.13\linewidth]{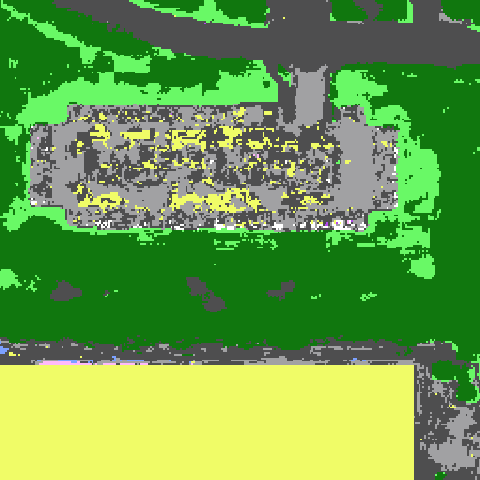}
}
\subfloat
{
\includegraphics[width=.13\linewidth]{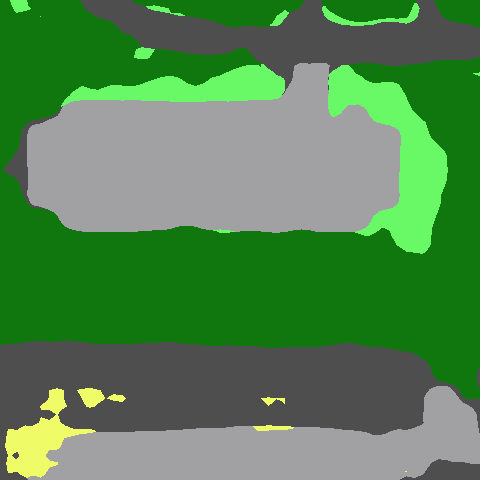}
}
\subfloat
{
\includegraphics[width=.13\linewidth]{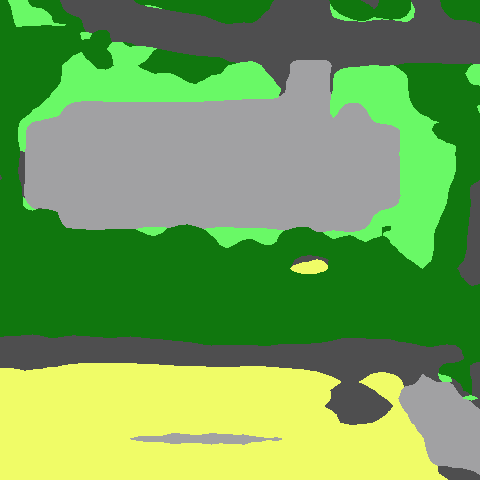}
}
\subfloat
{
\includegraphics[width=.13\linewidth]{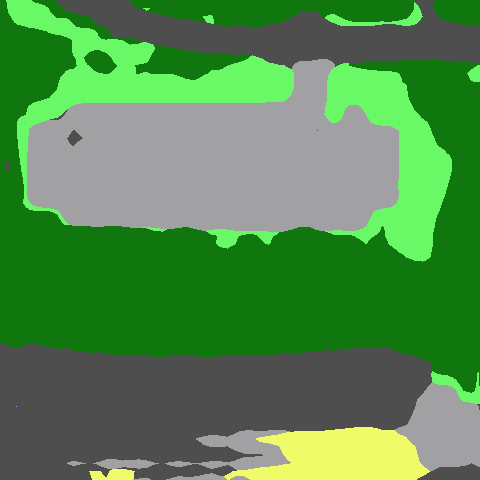}
}
\subfloat
{
\includegraphics[width=.13\linewidth]{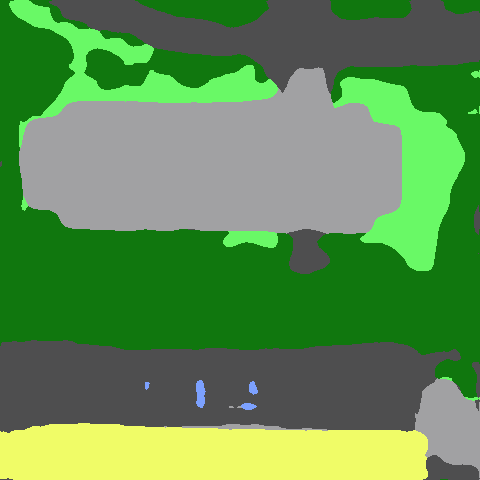}
}
\subfloat
{
\includegraphics[width=.13\linewidth]{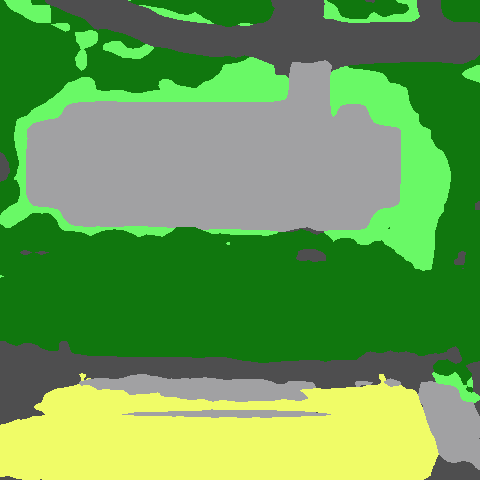}
}\\
\vspace{-0.7em}
\setcounter{subfigure}{0}
\subfloat[Image]
{
\includegraphics[width=.13\linewidth]{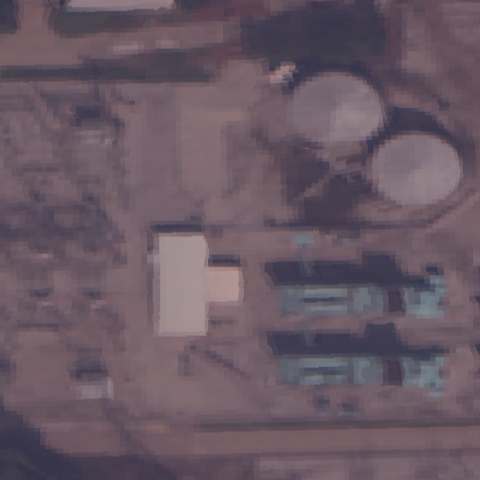}
}
\subfloat[Material Labels]
{
\includegraphics[width=.13\linewidth]{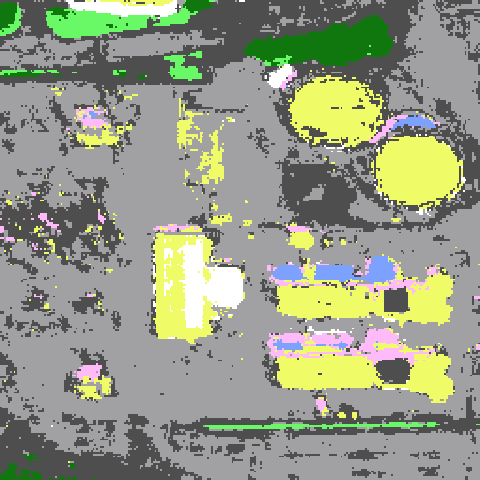}
}
\subfloat[FCN \cite{long2015fully}]
{
\includegraphics[width=.13\linewidth]{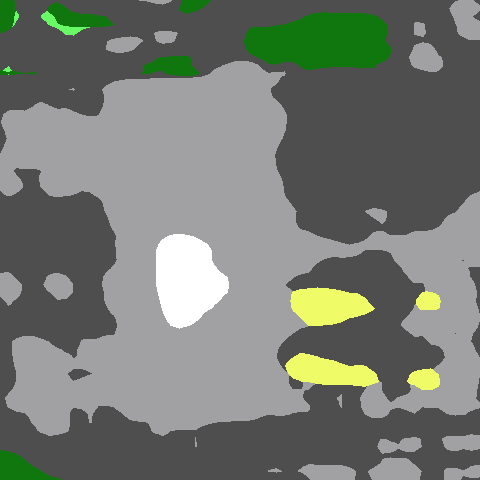}
}
\subfloat[PSPNet \cite{zhao2017pyramid}]
{
\includegraphics[width=.13\linewidth]{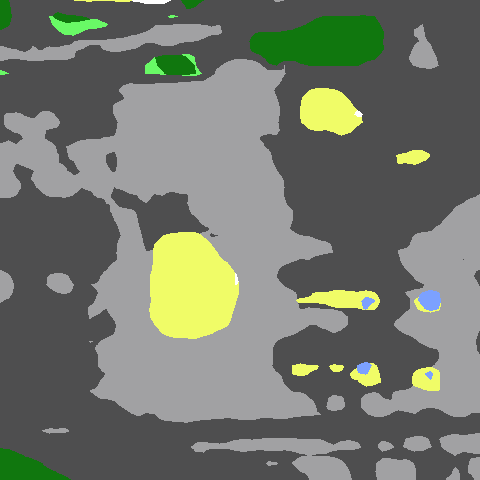}
}
\subfloat[EncNet \cite{zhang2018context}]
{
\includegraphics[width=.13\linewidth]{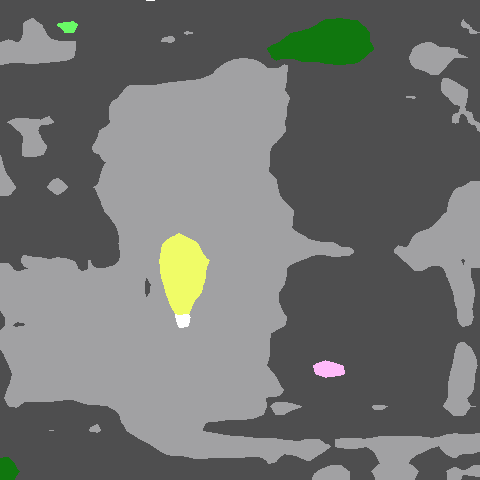}
}
\subfloat[MVCNet \cite{ma2017multi}]
{
\includegraphics[width=.13\linewidth]{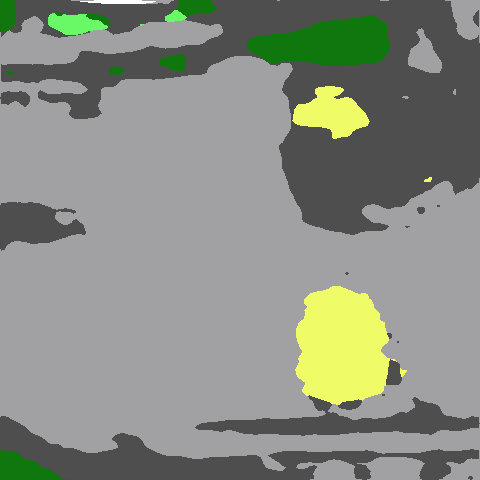}
}
\subfloat[AngLNet]
{
\includegraphics[width=.13\linewidth]{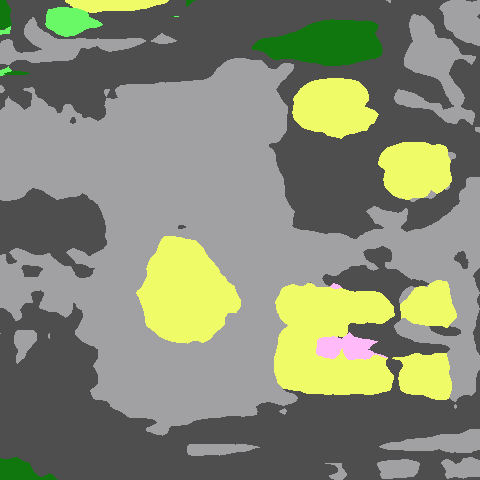}
}\\
\caption{Material segmentation results of AngLNet (g), dilated FCN and state-of-art semantic segmentation algorithms (c)-(f) on the satellite dataset. AngLNet improves the performance on both material prediction correctness and completeness as compared with the dataset material labels (b). 
}
\label{fig:result}
\vspace{-0.15in}
\end{figure*}

We evaluate and compare our AngLNet network on a material segmentation dataset derived from the SpaceNet challenge multispectral satellite image dataset~\cite{purri2019material}. The dataset contains multiple aligned and fully labeled images from three urban locations. The networks are trained on two regions and evaluated on the remaining region. The labeled satellite images are cropped into non-overlapping $500\times500$ images, resulting in 7421 images with 10 material classes.




The performance contribution of each addition to the baseline architecture is shown in Table~\ref{tab:ablation}.
Compared to the dilated FCN baseline, using the angular histogram improves the performance in Jacksonville from 72.6/26.6 to 73.9/28.5 in terms of pixel accuracy and mean IoU (\%). With stacking networks, the result further exceeds by 0.8/0.4 and reaches 74.7/28.9. We also experiment with a deeper network ( ResNet50 ) as a backbone network, but it does not consistently improve the segmentation performance.

\begin{table}[]
\centering
\begin{tabu}{c|c|c|c|c}
\tabucline[1pt]{-}
\multirow{2}{*}{Method} & \multicolumn{2}{c}{Jacksonville} & \multicolumn{2}{c}{Omaha} \\ 
 & pixACC & mIoU & pixACC & mIoU \\ \hline \hline
PSPNet \cite{zhao2017pyramid} & 73.5 & 28.0 & 67.1 & 29.9 \\ 
EncNet \cite{zhang2018context} & 73.7 & 28.1 & 67.2 & 29.2 \\ 
MVCNet \cite{ma2017multi} & 74.2 & 27.8 & 66.8 & 29.6 \\
AngLNet ({\small\TextRed{ours}}) &  \textbf{77.1} & \textbf{33.5} & \textbf{74.4} & \textbf{36.0} \\
\tabucline[1pt]{-}
\end{tabu}
\caption{
A comparison with state-of-the-art single view or multi-view segmentation algorithms. Notice that our method, AngLNet, includes multi-view fusion and achieves the best material segmentation.
}
\label{tab:state-of-art}
\vspace{-0.24in}
\end{table}

Table~\ref{tab:state-of-art} shows the segmentation result of AngLNet outperforming other single view (PSPNet \cite{zhao2017pyramid}, EncNet \cite{zhang2018context}) and multiview (MVCNet \cite{ma2017multi}) segmentation algorithms. All reported performances are based on pre-trained ResNet18, we use the same image augmentation for all the algorithms. To utilize multi-view information, each view is evaluated individually and the network outputs are combined via pixel-wise voting. We call this method \textit{multi-view fusion}. The weights for auxiliary loss and SE-loss in the PSPNet are both set to 0.2. We set the number of codewords in EncNet to 32. Since our dataset does not contain depth images, we do not use the depth view fusion branch for the MVCNet. 


The qualitative segmentation comparison of AngLNet, FCN and state-of-art semantic segmentation algorithms is shown in Figure~\ref{fig:result}. AngLNet improves the performance on both material prediction correctness and material prediction completeness. For example, in the first row, FCN predicts the metal building as half asphalt and half concrete, but  AngLNet classifies it correctly. 

\vspace{-0.15in}
\section{Conclusion}
\vspace{-0.08in}
Luminance histograms provide the statistics of natural images
and can be used as a strong material recognition cue. We revisit these classic concepts, incorporate modern deep learning networks and create a novel material recognition method for material recognition in satellite imagery.  The variation of local intensity with viewing angle is used to compute an angular luminance histogram. We show that utilizing this feature boosts  performance in modern deep learning architectures for material segmentation. 
\vspace{-0.15in}
\bibliographystyle{IEEEbib}
\bibliography{egbib}

\begin{thebibliography}{10}

\bibitem{dana1999reflectance}
Kristin~J Dana, Bram Van~Ginneken, Shree~K Nayar, and Jan~J Koenderink,
\newblock ``Reflectance and texture of real-world surfaces,''
\newblock {\em TOG}, 1999.

\bibitem{weyrich2006analysis}
Tim Weyrich, Wojciech Matusik, Hanspeter Pfister, and et~al.,
\newblock ``Analysis of human faces using a measurement-based skin reflectance
  model,''
\newblock in {\em ACM Transactions on Graphics (TOG)}. ACM, 2006, vol.~25, pp.
  1013--1024.

\bibitem{zhang2015reflectance}
Hang Zhang, Kristin Dana, and Ko~Nishino,
\newblock ``Reflectance hashing for material recognition,''
\newblock {\em CVPR}, pp. 3071--3080, 2015.

\bibitem{liu2014discriminative}
Chao Liu and Jinwei Gu,
\newblock ``Discriminative illumination: Per-pixel classification of raw
  materials based on optimal projections of spectral brdf,''
\newblock {\em IEEE transactions on pattern analysis and machine intelligence},
  vol. 36, no. 1, pp. 86--98, 2014.

\bibitem{dror2001estimating}
Ron~O Dror, Edward~H Adelson, and Alan~S Willsky,
\newblock ``Estimating surface reflectance properties from images under unknown
  illumination,''
\newblock in {\em HVEI}. International Society for Optics and Photonics, 2001,
  pp. 231--243.

\bibitem{motoyoshi2007image}
Isamu Motoyoshi, Shin'ya Nishida, Lavanya Sharan, and Edward~H Adelson,
\newblock ``Image statistics and the perception of surface qualities,''
\newblock {\em Nature}, p. 206, 2007.

\bibitem{sharan2008image}
Lavanya Sharan, Yuanzhen Li, Isamu Motoyoshi, Shin'ya Nishida, and Edward~H
  Adelson,
\newblock ``Image statistics for surface reflectance perception,''
\newblock {\em JOSA A}, vol. 25, no. 4, pp. 846--865, 2008.

\bibitem{zhu2016generative}
Jun-Yan Zhu, Philipp Kr{\"a}henb{\"u}hl, Eli Shechtman, and Alexei~A Efros,
\newblock ``Generative visual manipulation on the natural image manifold,''
\newblock in {\em European Conference on Computer Vision}. Springer, 2016, pp.
  597--613.

\bibitem{achanta2012slic}
Radhakrishna Achanta, Appu Shaji, Kevin Smith, Aurelien Lucchi, Pascal Fua,
  Sabine S{\"u}sstrunk, et~al.,
\newblock ``Slic superpixels compared to state-of-the-art superpixel methods,''
\newblock {\em PAMI}, pp. 2274--2282, 2012.

\bibitem{zhang2018context}
Hang Zhang, Kristin Dana, Jianping Shi, Zhongyue Zhang, Xiaogang Wang, Ambrish
  Tyagi, and Amit Agrawal,
\newblock ``Context encoding for semantic segmentation,''
\newblock in {\em CVPR}, 2018.

\bibitem{he2016deep}
Kaiming He, Xiangyu Zhang, Shaoqing Ren, and Jian Sun,
\newblock ``Deep residual learning for image recognition,''
\newblock in {\em CVPR}, 2016, pp. 770--778.

\bibitem{zeiler2014visualizing}
Matthew~D Zeiler and Rob Fergus,
\newblock ``Visualizing and understanding convolutional networks,''
\newblock in {\em ECCV}. Springer, 2014, pp. 818--833.

\bibitem{chen2017rethinking}
Liang-Chieh Chen, George Papandreou, Florian Schroff, and Hartwig Adam,
\newblock ``Rethinking atrous convolution for semantic image segmentation,''
\newblock {\em arXiv}, 2017.

\bibitem{ilg2017}
Eddy Ilg, Nikolaus Mayer, Tonmoy Saikia, Margret Keuper, Alexey Dosovitskiy,
  and Thomas Brox,
\newblock ``Flownet 2.0: Evolution of optical flow estimation with deep
  networks,''
\newblock in {\em CVPR}, 2017, vol.~2, p.~6.

\bibitem{zhang2017stackgan}
Han Zhang, Tao Xu, Hongsheng Li, Shaoting Zhang, Xiaolei Huang, Xiaogang Wang,
  and Dimitris Metaxas,
\newblock ``Stackgan: Text to photo-realistic image synthesis with stacked
  generative adversarial networks,''
\newblock {\em arXiv preprint}, 2017.

\bibitem{long2015fully}
Jonathan Long, Evan Shelhamer, and Trevor Darrell,
\newblock ``Fully convolutional networks for semantic segmentation,''
\newblock in {\em CVPR}, 2015, pp. 3431--3440.

\bibitem{zhao2017pyramid}
Hengshuang Zhao, Jianping Shi, Xiaojuan Qi, Xiaogang Wang, and Jiaya Jia,
\newblock ``Pyramid scene parsing network,''
\newblock in {\em CVPR}, 2017, pp. 2881--2890.

\bibitem{ma2017multi}
Lingni Ma, J{\"o}rg St{\"u}ckler, Christian Kerl, and Daniel Cremers,
\newblock ``Multi-view deep learning for consistent semantic mapping with rgb-d
  cameras,''
\newblock in {\em IROS}, 2017, pp. 598--605.

\bibitem{purri2019material}
Matthew Purri, Jia Xue, Kristin Dana, Matthew Leotta, Dan Lipsa, Zhixin Li,
  Bo~Xu, and Jie Shan,
\newblock ``Material segmentation of multi-view satellite imagery,''
\newblock {\em arXiv preprint arXiv:1904.08537}, 2019.

\end{thebibliography}

\end{document}